\pdfoutput=1
\documentclass[conference]{IEEEtran}
\IEEEoverridecommandlockouts
\usepackage{cite}
\usepackage{amsmath,amssymb,amsfonts}
\usepackage{algorithmic}
\usepackage{graphicx}
\usepackage{textcomp}
\usepackage{xcolor}

\usepackage{times}
\usepackage{soul}
\usepackage{url}
\usepackage[hidelinks]{hyperref}
\usepackage[utf8]{inputenc}
\usepackage[small]{caption}
\usepackage{graphicx}
\usepackage{amsmath}
\usepackage{amsthm}
\usepackage{booktabs}

\usepackage[switch]{lineno}

\usepackage{amssymb,amsfonts}
\usepackage{textcomp}
\usepackage{enumitem}
\usepackage{multirow}
\usepackage{array}

\usepackage{amsthm}

\usepackage{tikz}
\usepackage[margin=1in]{geometry}

\usepackage{makecell}

\def\BibTeX{{\rm B\kern-.05em{\sc i\kern-.025em b}\kern-.08em
    T\kern-.1667em\lower.7ex\hbox{E}\kern-.125emX}}
\begin{document}

\title{
Navigating the Black Box: Leveraging LLMs for Effective Text-Level Graph Injection Attacks
}

\author{\IEEEauthorblockN{1\textsuperscript{st} Given Name Surname}
\IEEEauthorblockA{\textit{dept. name of organization (of Aff.)} \\
\textit{name of organization (of Aff.)}\\
City, Country \\
email address or ORCID}
\and
\IEEEauthorblockN{2\textsuperscript{nd} Given Name Surname}
\IEEEauthorblockA{\textit{dept. name of organization (of Aff.)} \\
\textit{name of organization (of Aff.)}\\
City, Country \\
email address or ORCID}
\and
\IEEEauthorblockN{2\textsuperscript{nd} Given Name Surname}
\IEEEauthorblockA{\textit{dept. name of organization (of Aff.)} \\
\textit{name of organization (of Aff.)}\\
City, Country \\
email address or ORCID}
}
\author{
  Yuefei Lyu, 
  Chaozhuo Li, 
  Xi Zhang,
  Tianle Zhang\\
  \textit{Key Laboratory of Trustworthy Distributed Computing and Service (BUPT)} \\\textit{Ministry of Education,
  Beijing University of Posts and Telecommunications}, Beijing, China \\
}

\maketitle

\begin{abstract}

Text-attributed graphs (TAGs) integrate textual data with graph structures, providing valuable insights in applications such as social network analysis and recommendation systems.
Graph Neural Networks (GNNs) effectively capture both topological structure and textual information in TAGs but are vulnerable to adversarial attacks.
Existing graph injection attack (GIA) methods assume that attackers can directly manipulate the embedding layer, producing non-explainable node embeddings. Furthermore, the effectiveness of these attacks often relies on surrogate models with high training costs.
Thus, this paper introduces \textit{ATAG-LLM}, a novel black-box GIA framework tailored for TAGs.
Our approach leverages large language models (LLMs) to generate interpretable text-level node attributes directly, ensuring attacks remain feasible in real-world scenarios.
We design strategies for LLM prompting that balance exploration and reliability to guide text generation, and propose a similarity assessment method to evaluate attack text effectiveness in disrupting graph homophily.
This method efficiently perturbs the target node with minimal training costs in a strict black-box setting, ensuring a text-level graph injection attack for TAGs.
Experiments on real-world TAG datasets validate the superior performance of \textit{ATAG-LLM} compared to state-of-the-art embedding-level and text-level attack methods.

\end{abstract}


\section{Introduction}

Text-attributed graphs (TAGs), where nodes are associated with text attributes, effectively integrate textual data with graph structures. They are increasingly utilized in applications such as social network analysis, recommendation systems, and knowledge graphs \cite{tag1}. Graph Neural Networks (GNNs) play a crucial role in these contexts, capturing both topological dependencies and text semantics. However, GNNs are vulnerable to adversarial attacks \cite{GAD1,GAD2}. Unlike graphs devoid of textual content, TAGs face vulnerabilities in both their textual \cite{TAD-Textbugger,TAD-PETGEN,TAD-FastTextDodger} and structural \cite{GAD-survey} components. Addressing these vulnerabilities requires investigating adversarial attack methods specific to TAGs.

\begin{figure}
    \centering
    \includegraphics[width=\linewidth]{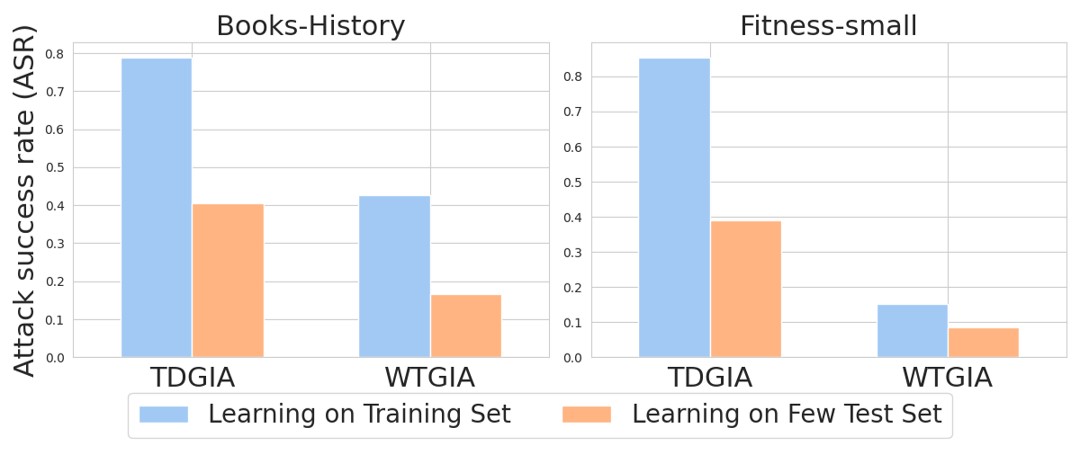}
    \caption{The attack success rate (ASR) decreases significantly when the surrogate model is learned on the different data from the target model. ASR is the ratio of the number of misclassified target nodes after attacking to all targeted nodes. Experiments are conducted in two datasets using TDGIA\cite{gia-tdgia} and WTGIA\cite{gia-text}, which are GIA methods depending on surrogate models.}
    \label{fig:intro}
\end{figure}

Currently, there is a shortage of adversarial attack methods specifically targeting TAGs. 
Existing approaches \cite{GAD1,GAD3,GAD2} often face challenges in real-world implementation.
Many methods attempt to deceive target models by deleting edges or modifying node attributes, typically falling outside the attacker’s capabilities due to the lack of necessary permissions to alter graph attributes. 
In practice, injecting new nodes is a more feasible attack strategy, known as graph injection attacks (GIAs). However, existing GIA methods \cite{gia-g2a2c,gia-tdgia,gia-hao,gia-camouflage,gia-single,gia-scalable} often generate node attributes at the embedding level rather than at the raw text level, assuming the attacker can directly attack the embedding layer, which is unrealistic in real-world scenarios. \cite{gia-text} proposed WTGIA method for TAGs, employing LLMs to improve embedding-level attack methods and achieve the text-level attacks, but it depends on surrogate models, which require significant costs to obtain training data and labels. As shown in Figure~\ref{fig:intro}, the performance of GIA methods using surrogate models is limited by their knowledge.

\begin{figure}
    \centering
    \includegraphics[width=\linewidth]{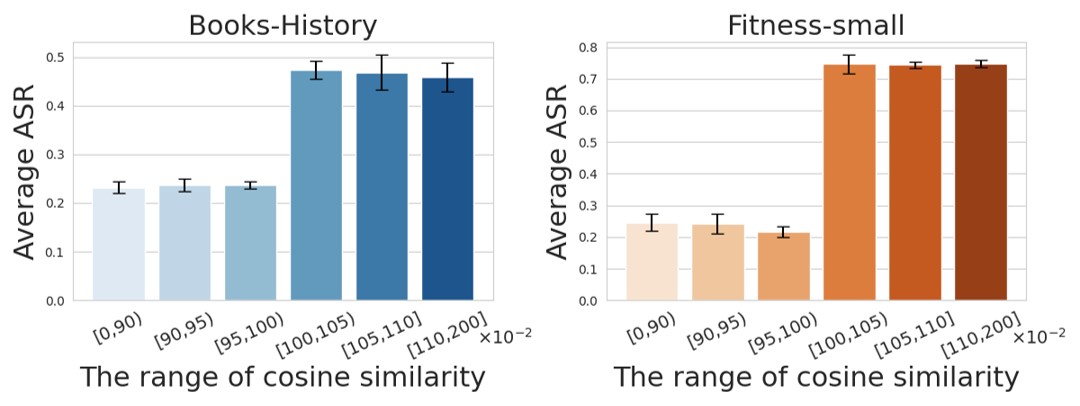}
    \caption{Experimental results of embedding-level random graph injection attack against EGNNGuard~\cite{gia-hao} in two datasets. The random GIA method generates random features for the injected nodes. Here we limit the cosine similarity between the target nodes and the injected nodes to a specific value range. We find that the attack success rate (ASR) decreases significantly once the similarity is lower than a certain value.}
    \label{fig:intro2}
\end{figure}

Based on these motivations, we focus on the text-level graph injection attack for TAGs in the black-box setting. On one hand, it is necessary to generate comprehensible node content at the text level. With the powerful text generation capabilities of large language models (LLMs), they naturally play a role in text-level GIAs. A straightforward approach is to drive LLMs to generate text dissimilar to the target node. However, excessively dissimilar nodes can disrupt graph homophily, making the attack more detectable \cite{gia-hao}, as shown in Figure~\ref{fig:intro2}. This issue is especially problematic in TAGs, where comprehensible text can help defenders identify irrelevant node injections. It requires text generation using LLMs with topic and category constraints. Previous research \cite{gia-text} has found that simple prompts do not effectively guide viable attack sample generation.
On the other hand, we consider a strict black-box scenario: the attacker can only access a limited number of node labels not used in training the target model, which are utilized for learning the attack model. During the attack phase, perturbations are applied to other nodes. 

Therefore, to execute an effective graph injection attack on TAGs, the following challenges must be addressed: 
(1) How can LLMs be guided to generate effective and inconspicuous text for GIAs? 
(2) How can attackers select optimal text and injection position for the injected nodes in a strict black-box setting?

In response, we propose a black-box graph injection attack framework, \textit{ATAG-LLM}, tailored for text-attributed graphs. 
This framework incorporates three strategies for LLM prompting, balancing exploration and reliability, to generate text that is distinct from the target node yet remains within the relevant topic and category. 
To select the most suitable text for the injected nodes from the generated candidates, we devise a similarity assessment method that evaluates text effectiveness, calculating similarity between the target nodes and the injected nodes in GIAs. This also aids in selecting LLM few-shot examples.
For determining injection positions, a graph explainer is utilized to identify important nodes to be connected.
Our approach avoids extensive model learning costs and reduces reliance on surrogate models.

\textbf{Contributions}:
(1)
We focus on novel text-level graph injection attacks with a realistic attack setting: conducting GIAs in a strict black-box scenario where node text content is directly generated rather than the non-explainable node embeddings. This approach ensures the attack is interpretable and applicable to real-world scenarios.
(2)
It proposes a novel LLM-based graph injection attack framework \textit{ATAG-LLM}. It designs various text generation strategies for LLM prompting and constructs the assessment method for text and injection position selection with low training cost.
(3)
By comparing our approach with state-of-the-art embedding-level and text-level graph injection attack methods on two real-world TAG datasets, we demonstrate the effectiveness of our approach across multiple scenarios.

\section{Background}

\subsection{Text-Attributed Graphs}
A text-attributed graph (TAG) is defined as $G = (\mathcal{V}, \mathcal{E}, \mathcal{S})$, where $\mathcal{V}$ is the set of nodes and $\mathcal{E}$ is the set of edges. Each node $v_i \in \mathcal{V}$ is associated with a sequential text feature $s_i \in \mathcal{S}$. The adjacency matrix is $\mathbf{A} \in \{0,1\}^{N \times N}$, where $N$ is the number of nodes. Using a text encoder, we represent each node by its corresponding text content, resulting in the feature matrix $\mathbf{X} \in \mathbb{R}^{N \times D}$, where $D$ is the feature dimension. For the node classification task, each node is associated with a label $y \in \{0, \ldots, C-1\}^N$. The goal is to predict the node labels in the testing set.

\subsection{Graph Neural Networks}
Given the adjacency matrix $\mathbf{A}$ and the feature matrix $\mathbf{X}$ of the graph $G$, the representation at the $l$-th layer of the Graph Neural Network (GNN) $f$ is described as:
\begin{equation}
    \mathbf{h}^l(i) = f^l \left( \mathbf{h}^{l-1}(i), \text{AGGR} \left( \{\mathbf{h}^{l-1}(j) : j \in \mathcal{N}_i\} \right) \right),
    \label{eq:gcl}
\end{equation}
where $\mathbf{h}^0(i) = \mathbf{X}[i]$, $\mathcal{N}_i$ is the neighborhood set of node $v_i$, AGGR is the aggregation function, and $f^l$ is a message-passing layer that processes the features of $v_i$ and its neighbors. The GNN model could output the label prediction result $\hat{y}_i$ of node $v_i$ as $\hat{y}_i=f(G,i)$.

\subsection{Graph Injection Attacks}
Given a clean graph $G$ and a target node $v_o \in \mathcal{T}$, the attacker injects $N_a$ nodes to form the perturbed graph $G' = (\mathcal{V}', \mathcal{E}', \mathcal{S}')$, thereby degrading the GNN's performance in predicting the label of $v_o$. The objective of a graph injection attack is formulated as:
\begin{equation}
\min_{G'} \mathbb{I}(f(G', v_o) = y_o | v_o \in \mathcal{T}),
\end{equation}
where $\mathbb{I}$ is an indicator function that returns the number of true conditions. The constraints for the graph injection attack are:
\begin{equation}
N_{a} \leq \Delta_N, 
\quad d_{a} \leq \Delta_d,
\end{equation}
where $d_a$ is the degree of the injected node $v_a$. $\Delta_N$ and $\Delta_d$ denote the maximum number of nodes that can be injected and the maximum degree of node $v_a$, respectively.

Here we define a strict black-box setting. The training and testing node sets are denoted as $\mathcal{V}_{train}$ and $\mathcal{V}_{test}$, respectively. The target model $f$ is trained on $\mathcal{V}_{train}$. Attackers have access only to the ground truth of nodes in $\mathcal{V}'$, a small subset of $\mathcal{V}_{test}$, which they use to learn a surrogate model. The goal of the graph injection attack is to perturb the target nodes in another subset $\mathcal{T}$, ensuring that $\mathcal{T} \cap \mathcal{V}' = \emptyset$.

\section{Methodology}

\begin{figure*}
    \small
    \centering
    \includegraphics[width=0.8\linewidth]{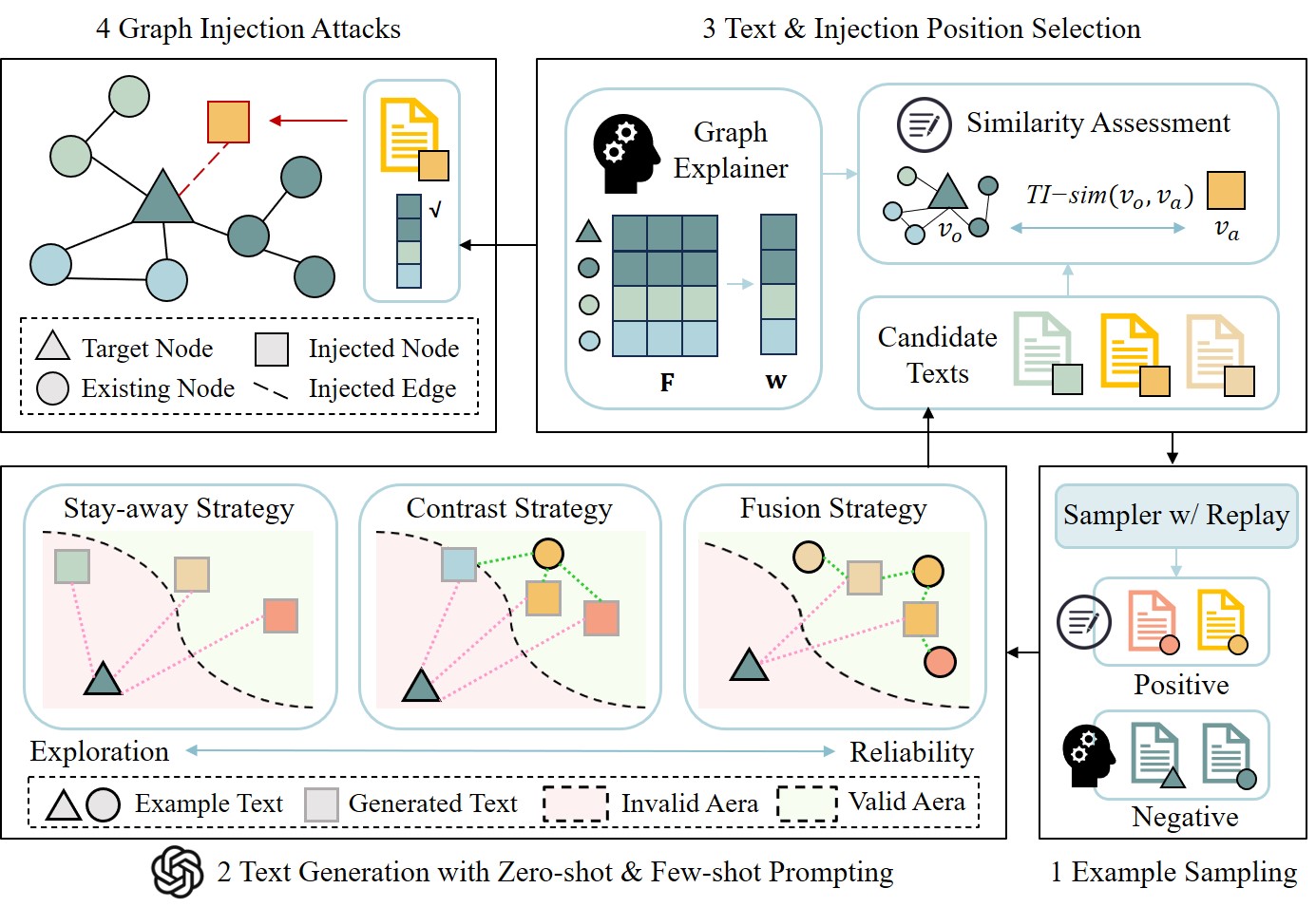}
    \caption{The framework of our method \textit{ATAG-LLM}, comprising four stages: example sampling, text generation, text and injection position selection, and graph injection attack.
    In the text generation stage, we visualize the semantic space as planes depicted in the figure, illustrating the motivation behind the three strategies.  
    The black dashed line represents the effectiveness boundary for the injected node text. If the generated text for the injected node lies above this dashed line, it indicates a successful attack.   The pink and green dashed lines respectively denote the intentions of semantic distancing and proximity.}
    \label{fig:framework}
\end{figure*}

The attack framework of \textit{ATAG-LLM} is illustrated in Figure \ref{fig:framework}. To execute a graph injection attack on text-attributed graphs, 
the key is to appropriately disrupt graph homophily by injecting nodes dissimilar to the target node and its neighbors, while remaining unnoticeable~\cite{gia-hao}.
Initially, our approach provides positive and negative examples to a general LLM, and drives it to generate the candidate texts for the injected node with three strategies.
Subsequently, it constructs a similarity assessment method to select the injected node text, which also supports example sampling in the first step. We employ a graph explanation model to determine node importance, which is used for both similarity assessment and injection position selection. Finally, with the selected text and positions, the graph injection attack is executed.

\subsection{Similarity Assessment Method}

\begin{figure}
    \centering
    \includegraphics[width=\linewidth]{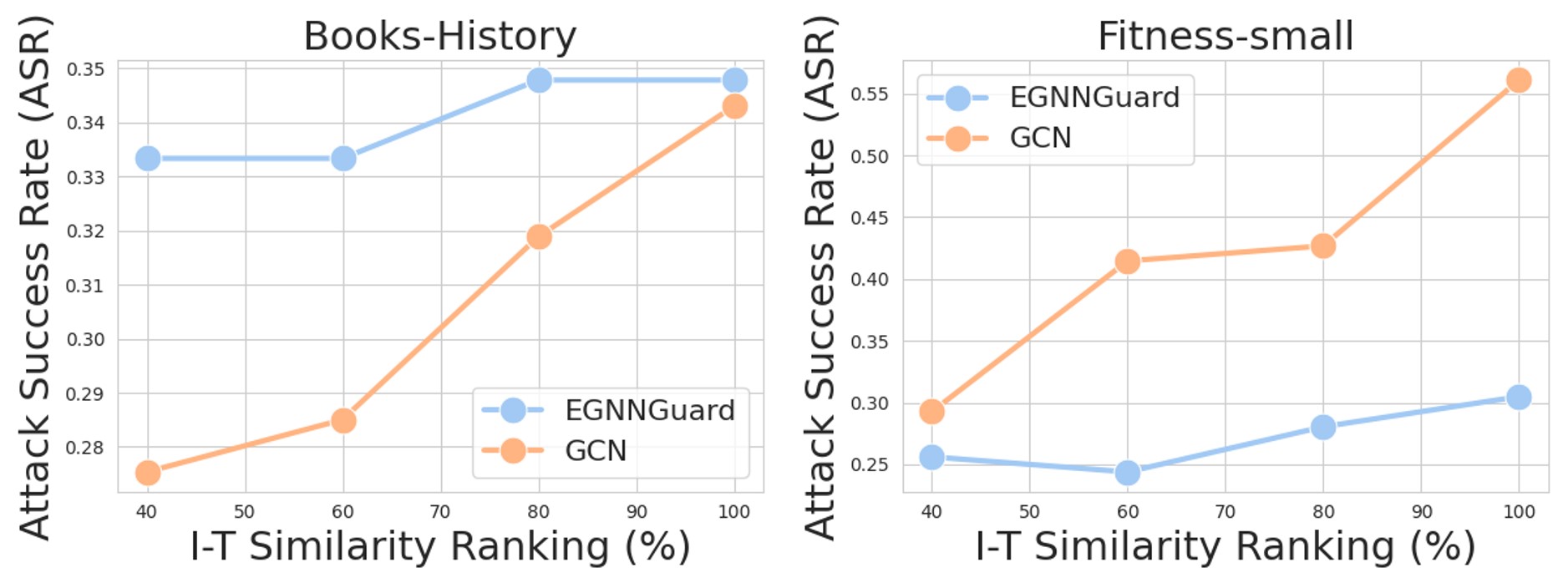}
    \caption{Preliminary GIA experiments against GCN and EGNNGuard using two datasets. Existing nodes are randomly sampled, and the T-I Similarities is calculated and ranked. For a given \( r \), we select the existing node whose T-I similarity falls within the top \( r\% \) of the ranking and copy its text content to the injected node. ASR increases as T-I similarity decreases.}
    \label{fig:intro3}
\end{figure}

Numerous studies \cite{heter1, gia-hao} have identified that one of the most effective strategies for attacking GNNs is disrupting the graph homophily distribution. Homophily indicates a tendency for similar nodes to be connected. A straightforward approach is to inject new nodes that are as dissimilar as possible from the target nodes. In response, some studies have proposed defense mechanisms, such as EGNNGuard \cite{gia-hao}, to restore the original homophily and limit disruption. Consequently, to achieve effective and imperceptible graph injection attacks, we propose a similarity assessment method to quantify the dissimilarity between injected and target nodes, ensuring appropriate disruption of the graph's homophily.

For two single nodes \(v_i\) and \(v_j\), given their initial features \(\mathbf{x}_i\) and \(\mathbf{x}_j\), the semantic gap can be measured using cosine similarity:
\begin{equation}
c(i,j)=\textit{cos\_sim}(v_i,v_j) = \frac{\mathbf{x}_i \cdot \mathbf{x}_j}{\|\mathbf{x}_i\| \|\mathbf{x}_j\|}.
\end{equation}
However, for graph structured data, the topology structure should be considered. GNNs enhance the ability of node classification by propagating information among connected nodes, aggregating information from their $k$-hop neighborhood. 
To effectively assess similarity for attacking GNNs, it is beneficial to consider the influence of nodes within the $k$-hop neighborhood of the target node on GNNs' predictive outcomes, referred to as \textit{node importance}.

Thus, for graph injection attacks on GNNs, we propose a specialized node similarity measure as follows. 
Given a fixed target node $v_o$, we define \textit{Target-Injected Node Similarity (T-I Similarity)} of an injected node $v_a$ as
\begin{equation}
\textit{TI-sim}(v_o,v_a) = (1-\lambda)c(o,a) + \lambda\sum_{v_i \in \mathcal{N}_o^k}{w_i c(i,a)},
\end{equation}
where \(\mathcal{N}_o^k\) represents the $k$-hop neighborhood of the target node $v_o$, and \(\lambda\) is the trade-off coefficient. $w_i$ is the node importance of $v_i$.
In subsequent discussions, all T-I Similarity values are calculated relative to the pre-specified target node $v_o$.

We utilize GNNExplainer~\cite{gnnex} to obtain the node importance vector \(\mathbf{w}=\{w_i\}\). Specifically,
\begin{equation}
\mathbf{F} = \text{GNNExplainer}(g(G))
\end{equation}
\begin{equation}
\mathbf{w} = \sigma_m( \sum_{j=1}^{D} \mathbf{F}_{:,j})
\label{eq:w}
\end{equation}
where \(\mathbf{F} \in \mathbb{R}^{N \times D}\) is the node feature importance matrix optimized jointly with graph structure and node features, and \(g\) is a GNN surrogate model trained using labels from \(\mathcal{V}'\). $\sigma_m()$ is the truncation and normalization function to retain the top $m$ most important nodes.

With this similarity assessment method, we can analyze the relationship between attack effectiveness and T-I similarity to identify optimal attack strategies. Preliminary experiments, as shown in Figure~\ref{fig:intro3}, randomly sample text content from existing nodes to serve as injected node text content. We observed that lower T-I similarity correlates with better attack performance against the defensive GNNs. This suggests that even if the injected node is semantically dissimilar to the target node, as long as it fits the context topic and appears natural, it is difficult for the target model to detect it. Large language models can effectively meet the constraints of topic and text naturalness, leaving the challenge of generating text with low T-I similarity under these constraints.

\subsection{Text Generation and Injection Position Selection}

\noindent\textbf{LLMs Prompt Design}.
Semantic similarity in natural language involves complex and nuanced relationships that are often challenging to quantify and define precisely. Without fine-tuning, aligning our defined T-I similarity with an LLM's understanding of semantic similarity is challenging. However, fine-tuning LLMs is computationally expensive and difficult to generalize across different datasets. Therefore, we prefer to enhance the effectiveness of text generation by designing effective prompts for LLMs.
Specifically, we have developed three strategies:

\textit{Stay-Away Strategy}: This zero-shot prompting requires generating text that is as dissimilar as possible from the target node text.

\textit{Contrast Strategy}: This few-shot prompting uses the target node text and texts from highly important neighboring nodes as negative examples, while existing node texts in the dataset with low T-I similarity are used as positive examples. The generated text should be dissimilar to the negative examples and similar to the positive examples.

\textit{Fusion Strategy}: Similar to the contrast strategy, but the LLM is instructed to semantically integrate the provided positive examples,  and it is required to include keywords from the positive examples while excluding those from the negative examples.

These strategies progressively increase in constraint, reducing exploration range but enhancing reliability. As illustrated in the second stage of Figure~\ref{fig:framework}, without positive sample guidance, the stay-away strategy may explore various directions, presenting both risks and opportunities. The fusion strategy, under multiple constraints, increases the likelihood of generating text within a safe zone in the semantic space, although it may lead to local optima. The contrast strategy balances exploration and reliability. Detailed prompts are shown in Figure~\ref{fig:prompt} and Figure~\ref{fig:prompt2}. The system prompts define the text generation strategy and specify requirements for topics and categories. User inputs include positive and negative examples along with their T-I similarity.

\begin{figure}
    \centering
\begin{tikzpicture}
\node[draw, dashed, rectangle, fill=blue!5, text=black, text width=0.91\linewidth, align=left] at (0,0) {
\textit{Stay-away}:\\ Your task is to craft a book description and title that remain distinctly \textbf{different from} the user's input negative examples in terms of length, vocabulary, theme, and style.};

\node[draw, dashed, rectangle, fill=blue!5, text=black, text width=0.91\linewidth, align=left, yshift=-2.6cm] at (0,0) {
\textit{Contrast}:\\ Your task is to craft a book description and title that closely \textbf{mirror} the user's input positive examples in terms of length, vocabulary, theme, and style, while ensuring they remain distinctly \textbf{different from} the user's input negative examples.};

\node[draw, dashed, rectangle, fill=blue!5, text=black, text width=0.91\linewidth, align=left, yshift=-6.05cm] at (0,0) {
\textit{Fusion}:\\ Your task is to craft a book description and title that \textbf{naturally weave together the semantics} of the user's input positive examples, reflecting their length, vocabulary, theme, and style. Ensure the content remains distinctly different from the user's input negative examples by \textbf{incorporating keywords} from the positive examples and \textbf{avoiding} those from the negative examples.
};

\node[draw, dashed, rectangle, fill=blue!5, text=black, text width=0.91\linewidth, align=left, yshift=-9.1cm] at (0,0) {
\textit{General}:\\
The book should fit within the \textbf{History} genre and fall into one of these categories: \{\textcolor{blue}{names\_of\_categories}\}.
};
\end{tikzpicture}
    \caption{The system prompts of three text generation strategies for the Books-History dataset.}
    \label{fig:prompt}
\end{figure}

For each strategy, $N_r$ rounds of text generation are performed to produce $3N_r$ candidate texts, from which the text with the lowest T-I similarity is selected as the injected node text.

\noindent\textbf{Few-Shot Example Sampling}. 
Providing effective positive and negative examples to LLMs is crucial for text generation. 
As mentioned earlier, the selection of negative examples depends on node importance $\mathbf{w}$. Excluding the target node, we select the top $N_n$ most important node texts as negative examples. 
For positive examples, a sampler with a replay buffer is employed, using T-I similarity to rank and filter examples. 
We use a simple random sampler, which can be further enhanced to a neural network-based sampling network.
Each target node $v_o$ has a replay buffer to store the optimal positive example group for guiding text generation.
Specifically, for each injected node $v_a$, there are $N_r$ rounds of text generation.
In each round, the texts of $N_e$ existing nodes are sampled from the dataset, and the T-I similarities are calculated. 
The $N_p$ examples with the lowest T-I similarity values are selected as few-shot examples and provided to the LLM for text generation.
We assess the effectiveness of the selected positive examples as few-shot examples based on the T-I similarity of the generated candidate node.
The positive example group $\{v_p\}$ which yields the candidate node text the lowest T-I similarity is stored in the replay buffer and used in the final round of positive example sampling to ensure baseline effectiveness in guiding text generation.

\noindent\textbf{Injection Position Selection}.
We utilize node importance $\mathbf{w}$ calculated by Eq.~(\ref{eq:w}) to rank and select the nodes to be connected, deciding the injection position. 
Generally, the first connected node is the target node. We consider two scenarios: direct attack and indirect attack. The former allows the node to be injected directly into the target node, while the latter does not permit this connection. The indirect attack enhances the imperceptibility of the perturbation or corresponds to scenarios where the target node cannot be directly attacked. 

\begin{figure}
    \centering
    \begin{tikzpicture}
    \node[draw, dashed, rectangle, fill=blue!5, text=black, text width=0.865\linewidth, align=left] at (0,0) {
    Here are examples of high and low semantic similarity to guide your task.\\
    Negative Example 1: \{\textcolor{blue}{target\_node\_text}\}\\
    Negative Example 2: \{\textcolor{blue}{neg\_ex\_text1}\}\\
    The similarity between Negative Example 1 and Negative example 2 is \{\textcolor{blue}{neg\_ex\_tisim}\}.\\
    Positive Example 1: \{\textcolor{blue}{pos\_ex\_text1}\}.\\
    The similarity between Negative Example 1 and Positive Example 1 is \{\textcolor{blue}{pos\_ex\_tisim}\}.\\
    ...\\
    Please generate the new book and description.\\
    Limit length: \{\textcolor{blue}{limit\_max\_length}\} words.
    };
    \end{tikzpicture}
    \caption{The user prompt for the Books-History dataset.}
    \label{fig:prompt2}
\end{figure}

\section{Experiments}

\begin{table}
    \centering
    \caption{Dataset statistics. Avg. and Std. denote the average value and standard deviation, respectively. The maximum text length for LLM text generation is set to "Avg. + 2Std."}
    \label{tab:dataset}
    \begin{tabular}{l|rr}
    \toprule
        ~ & Books-History & Fitness-small \\ \midrule
        Nodes & 41,551 & 8,886 \\ 
        Edges & 358,574 & 36,582 \\ 
        Classes & 12 & 13 \\ 
        Avg. Text Length & 227.81 & 21.53 \\ 
        Std. Text Length & 308.08 & 5.85 \\  \bottomrule
    \end{tabular}
\end{table}

\subsection{Dataset}
The experiments are conducted on two text-attributed graph datasets: \textit{Books-History} \cite{tag1} and \textit{Fitness-small} (a subset of \textit{Sports-Fitness} \cite{tag1}). The \textit{Books-History} dataset is extracted from the Amazon-Books dataset, focusing on books categorized as 'History'. Nodes represent books, and edges indicate frequent co-purchases or co-views. Each node is associated with a book's title and description. The \textit{Sports-Fitness} dataset is derived from the Amazon-Sports dataset and includes items related to 'Fitness'. Nodes represent Amazon items attributed with item titles, and edges indicate frequent co-purchases or co-views. We randomly sample 5\% of nodes, retaining the original label distribution to construct the \textit{Fitness-small} dataset. Both \textit{Books-History} and \textit{Fitness-small} are split with a ratio of 6:2:2 for training, validation, and testing. Dataset statistics are provided in Table \ref{tab:dataset}.

\subsection{Baselines}
We compare our method to four graph injection attack methods, including embedding-level GIAs (\textit{TDGIA}, \textit{G2A2C}) and text-level GIAs (\textit{RandLLM}, \textit{WTGIA}).
\begin{itemize}
\item \textit{TDGIA} \cite{gia-tdgia}: A GIA method that combines topological defective edge selections with smooth feature generation. It relies on a surrogate model.
\item \textit{G2A2C} \cite{gia-g2a2c}: An RL-based GIA method for black-box attacks, modeling node injection attacks as a Markov decision process.
\item \textit{RandLLM}: It generates text randomly using LLMs for injected node content with a specific topic and limited text length, and connects injected nodes to the target node directly or to a neighbor randomly.
\item \textit{WTGIA} \cite{gia-text}: It extends embedding-level GIA to the text level with word-frequency-based embeddings and a row-wise constrained FGSM algorithm based on \cite{gia-scalable}. LLMs are employed to solve the "word-to-text" task, implemented based on TGDIA.
\end{itemize}
The uniform constraint is the limited knowledge of labels for surrogate models with transductive learning.

\subsection{Target Models and Target Nodes}
We utilize a 2-layer graph convolutional network (GCN)~\cite{gcn} and a 3-layer EGNNGuard \cite{gia-hao} as the target models for our experiments. The latter is a robust GNN model designed to defend against obvious homophily destruction. The performance of the target models on a clean graph is shown in Table~\ref{tab:gnn}. The target node set $\mathcal{T}$ is sampled from nodes that are correctly classified in the testing set, with a 1\% and 5\% ratio for Books-History and Fitness-small, respectively. We select 25\% of nodes with the highest confidence margins, 25\% with the lowest confidence margins, and randomly sample 50\% from the middle range of prediction margins.

\subsection{Experimental Setup}
We pre-train a 2-layer GCN model as the surrogate model $g$. It is trained using a semi-supervised method over 1000 epochs with 10\% true labels on the testing set, and samples with classification probabilities exceeding 0.75 are given pseudo-labels. The pre-trained \textit{bert-base-uncased} model is employed to encode node text content. The surrogate and target GNN hidden dimension is 128. All networks are trained with the Adam optimizer at a learning rate of 0.01. The trade-off coefficient $\lambda$ for T-I similarity is 0.1. The truncation coefficient $m$ is 10. The LLM for text generation is \textit{gpt-4o-mini}. We sample few-shot examples with $N_n=1$, $N_p=2$, $N_r=2$, and $N_e=1000$. There are 10 groups of target nodes for repeated experiments. Experiments are conducted using \textit{NVIDIA vGPU-32GB} on \textit{Ubuntu 20.04}, and the framework is implemented with Python and PyTorch.

\begin{table*}[!ht]
    \centering
    \caption{The attack success rate (\%) against GCN and EGNNGuard on two datasets. Higher ASR indicates better attack performance. 
    ELGIA and TLGIA indicates embedding-level GIA and text-level GIA, respectively.
    \textbf{Boldface font} and $\ast$ denotes the best performance among text-level GIA methods and all methods, respectively.}
    \label{tab:att-gcn}
    \begin{tabular}{c|c|c|rrr|rrr}
    \toprule\midrule
    ~ &~ & ~ & \multicolumn{3}{c|}{Books-History} & \multicolumn{3}{c}{Fitness-small} \\ \midrule
    ~ & ~ & $\Delta_N$ & \multicolumn{1}{c}{5} & \multicolumn{1}{c}{10} & \multicolumn{1}{c|}{20} & \multicolumn{1}{c}{5} & \multicolumn{1}{c}{10} & \multicolumn{1}{c}{20} \\ \midrule\midrule
    \multirow{5}{*}{GCN} & \multirow{2}{*}{ELGIA} & TDGIA & $\ast$40.43±2.64 & $\ast$42.71±2.16 & $\ast$42.71±1.35 & 39.02±8.31 & 40.86±8.65 & 41.34±7.70\\ 
    & ~ & G2A2C & 23.67±0.68 & 30.44±1.18 & 33.33±3.13 & 50.41±2.07 & 59.35±2.51 & 69.92±4.15 \\ \cmidrule(lr){2-9}
    & \multirow{3}{*}{TLGIA} & RandLLM & \textbf{25.60±3.88} & 30.92±3.90 & \textbf{36.72±4.48} & 32.93±3.59 & 37.40±1.52 & 42.28±3.50 \\ 
    & ~ & WTGIA & 16.57±1.14 & 19.57±1.57 & 20.00±1.28 & 8.54±3.70 & 10.37±3.33 & 10.37±3.59\\ 
    & ~ & ATAG-LLM & 25.12±3.62 & \textbf{33.33±3.13} & 36.23±3.13 & $\ast$\textbf{73.17±1.73} & $\ast$\textbf{85.77±2.07} & $\ast$\textbf{91.06±0.58} \\ \midrule\midrule
    \multirow{5}{*}{EGNNGuard} & \multirow{2}{*}{ELGIA} & TDGIA & 16.71±1.81 & 18.29±1.07 & 17.14±1.69 & 5.12±1.06 & 4.76±1.01 & 5.85±0.91\\ 
    & ~ & G2A2C & 26.57±1.37 & 27.54±1.18 & 30.92±0.68 & 23.17±2.99 & 29.68±4.02 & 34.15±2.63 \\ \cmidrule(lr){2-9}
    & \multirow{3}{*}{TLGIA} & RandLLM & 28.99±1.18 & 39.99±1.81 & 43.00±3.80 & 49.59±3.77 & 63.01±1.52 & 78.46±2.07 \\ 
    & ~ & WTGIA & 14.43±1.49 & 13.86±1.12 & 14.00±1.07 & 4.02±1.64 & 3.17±0.98 & 4.02±1.34\\ 
    & ~ & ATAG-LLM & $\ast$\textbf{33.33±3.13} & $\ast$\textbf{40.10±4.93} & $\ast$\textbf{51.21±1.37} & $\ast$\textbf{78.86±2.87} & $\ast$\textbf{91.46±2.99} & $\ast$\textbf{93.90±1.99} \\ \bottomrule
    \end{tabular}
\end{table*}


\subsection{Attack Performance}
The attack performance is measured by the attack success rate (ASR) as
\begin{equation}
    ASR= \frac{\text{\# misclassified target nodes in $\mathcal{T}$}}{|\mathcal{T}|}.
\end{equation}
The edge budget $\Delta_d=1$ allows each injected node to be added with a single link. It is worth noting that, as shown in \cite{gia-text}, due to the trade-off between performance and interpretability, text-level GIAs often struggle to exceed the effectiveness of embedding-level GIAs. Without considering whether the injected node embeddings correspond to interpretable text content, the optional embedding space of embedding-level GIAs is much larger than that of text-level GIAs.

The attack performance against GCN and EGNNGuard are presented in Table~\ref{tab:att-gcn}. 
For targeting the GCN model, our method achieves optimal or near-optimal results among text-level GIAs, significantly outperforming the embedding-level methods G2A2C and TDGIA on the Fitness-small dataset. 
RandLLM also shows strong attack performance, particularly on the Books-History dataset. This may be attributed to the less stringent constraints on generated text length in Books-History, as detailed in Table \ref{tab:dataset}, which allows LLMs to explore semantic spaces distant from the target node. Conversely, RandLLM does not perform as well on the Fitness-small dataset, where text length is more restricted. This highlights the effectiveness of our text generation strategy under conditions with constrained maximum text length.

As for the robust model EGNNGuard, our method achieves optimal performance in all scenarios. When the target model shifts from GCN to EGNNGuard, the performance of most other attack methods noticeably decreases, whereas the effectiveness of our method improves. This suggests that our approach exhibits superior imperceptibility and adaptability when confronting defensive GNNs, underscoring its robustness in more challenging environments.

The results for indirect attacks are presented in Table \ref{tab:att-in}. Since TDGIA does not support indirect attacks, comparisons are made only with the RandLLM and G2A2C methods. Our method consistently achieves optimal or near-optimal performance, indicating that GNNExplainer effectively selects the injection position even when the surrogate model's training is limited.

\begin{table}[!ht]
    \centering
    \caption{The attack success rate (\%) against GCN with the limit of indirect attack. \textbf{Boldface font} denotes the best performance among all methods.}
    \label{tab:att-in}
    \begin{tabular}{c|c|cc}
        \toprule
        Dataset & Method & $\Delta_N=5$ & $\Delta_N=10$  \\ \midrule
        \multirow{3}{*}{Books-History}  & RandLLM & 18.36±0.68 & 20.29±1.18  \\ 
         & G2A2C & 18.05±0.91 & \textbf{21.74±1.06}  \\ 
         & ATAG-LLM & \textbf{19.32±2.46} & 20.29±2.05  \\ \midrule
        \multirow{3}{*}{Fitness-small} & RandLLM & 21.55±1.52 & 28.05±3.98  \\ 
         & G2A2C & 28.05±4.34 & 33.74±3.77  \\ 
         & ATAG-LLM & \textbf{32.11±2.51} & \textbf{43.50±2.30}  \\ \bottomrule
    \end{tabular}
\end{table}

\begin{table}[!ht]
    \centering
    \caption{The accuracy (\%) of GCN and EGNNGuard on validation and testing sets.}
    \label{tab:gnn}
    \begin{tabular}{l|cc|cc}
    \toprule
        ~ & \multicolumn{2}{c|}{Books-History} & \multicolumn{2}{c}{Fitness-small} \\ \midrule
        ~ & Validation & Testing & Validation & Testing  \\ \midrule
        GCN & 84.92 & 83.50 & 90.27 & 89.10  \\ 
        EGNNGuard & 84.18 & 83.24 & 89.36 & 87.05  \\ \bottomrule
    \end{tabular}
\end{table}

\subsection{Analysis of Text Generation Strategies}
\begin{figure}
    \centering
    \includegraphics[width=\linewidth]{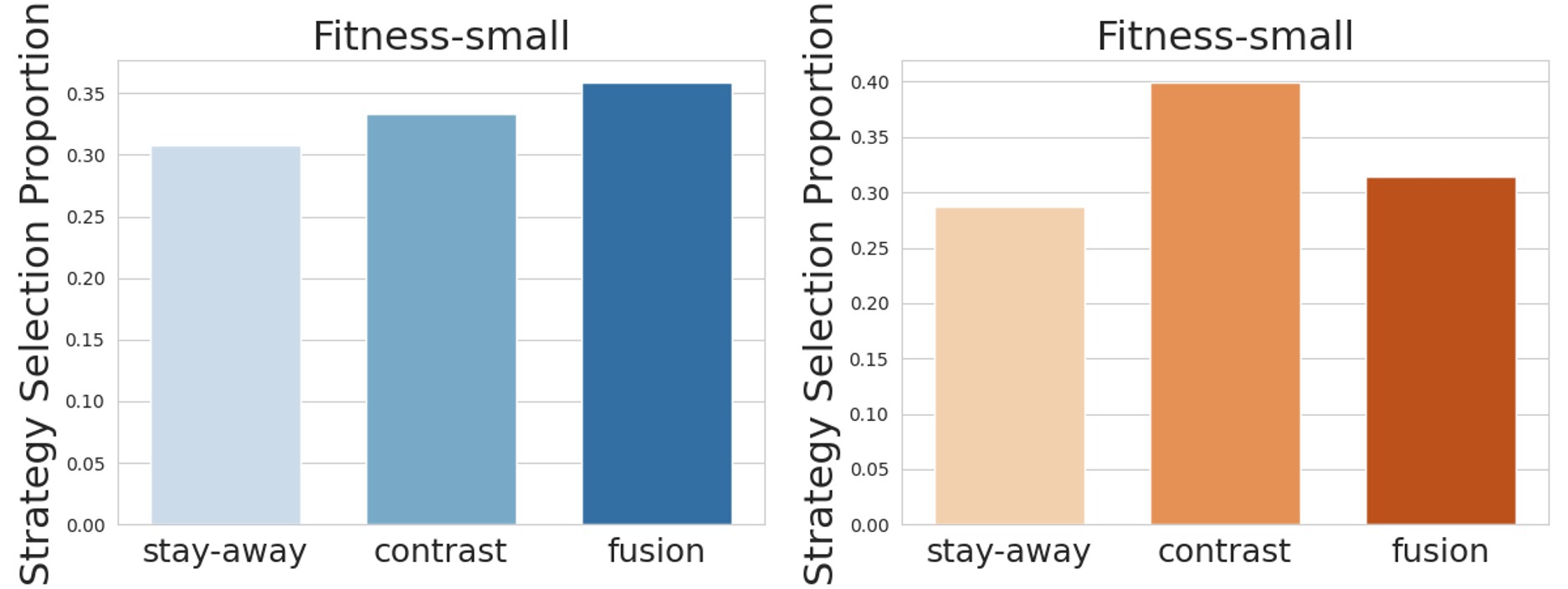}
    \caption{The proportion of strategies adopted by successful injection attacks in two datasets. }
    \label{fig:s1}
\end{figure}

\begin{figure}
    \centering
    \includegraphics[width=\linewidth]{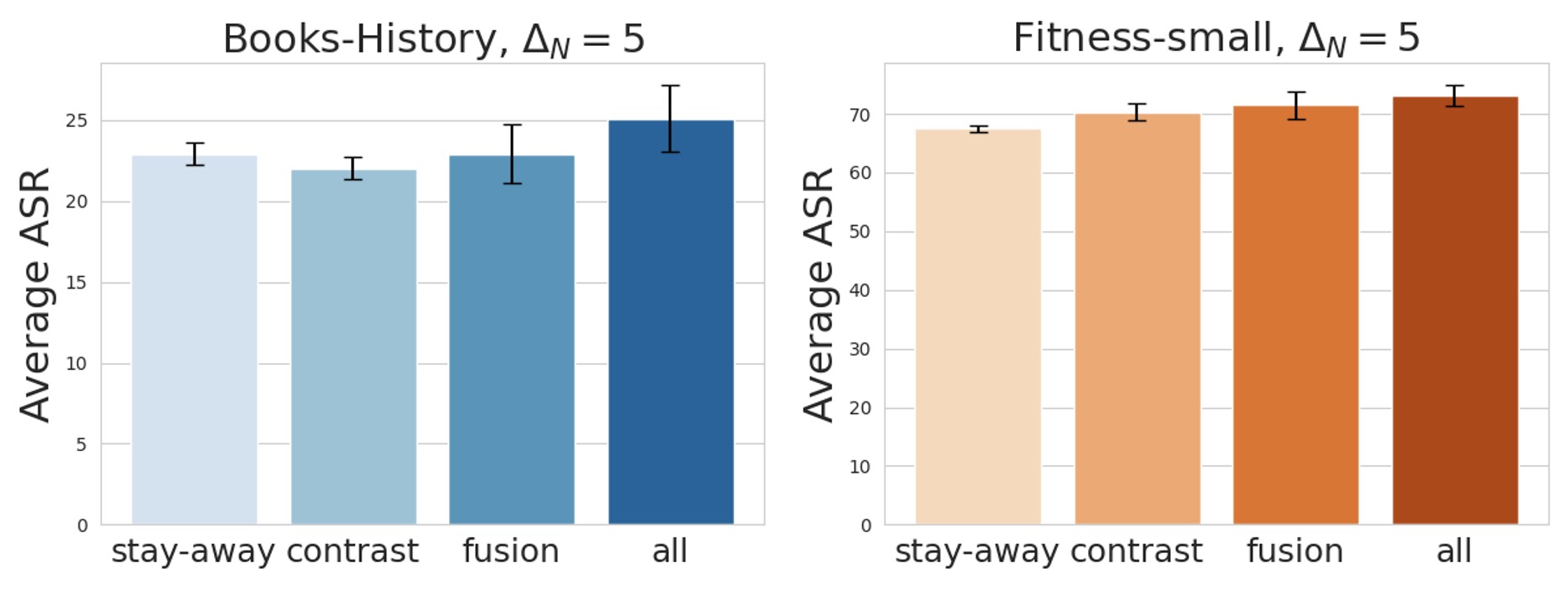}
    \caption{The attack success rate (\%) of employing single text generation strategy and all strategies.}
    \label{fig:s2}
\end{figure}

Figure \ref{fig:s1} presents the statistics of strategy selection, revealing a significant disparity in strategy preferences across different datasets. Figure \ref{fig:s2} illustrates the attack success rates for employing single text generation strategies compared to all strategies combined, indicating that the combined use of three strategies is superior to any single strategy. This underscores the necessity of designing multiple strategies to enhance the generalizability and effectiveness of our attack methods.

\section{Related Work}
\textbf{Graph adversarial attacks}. It can be broadly categorized into two main types: graph modification attacks (GMAs) and graph injection attacks (GIAs). GMAs \cite{GAD1,GAD2} involve perturbing the existing graph structure or node features, whereas GIAs focus on injecting new nodes into the graph. Among these, black-box GIAs \cite{gia-nipa,gia-g2a2c} are considered more practical as they do not require any knowledge of the target model. In contrast, white-box GIAs \cite{gia-single,gia-tdgia}, which typically rely on model gradients, can be adapted for black-box settings using a surrogate model. In the context of text-attributed graphs, \cite{gia-text} introduces three methods for achieving text-level GIAs, highlighting the trade-off between attack performance and text interpretability. \cite{gia-mia} focuses on rumor detection tasks, where it injects messages into message propagation trees to perform the attack.

\textbf{Graph model robustness and LLMs}. Recently, LLMs have increasingly been applied to graph-related tasks,
primarily in three directions: LLM as Enhancer, LLM as Predictor, and GNN-LLM Alignment \cite{gllm-survey}. 
Some studies have focused on the integration of LLMs with graph model robustness.
\cite{gllm-robust} explores the robustness of LLMs as enhancers and predictors in the context of structural and textual perturbations in graph data.
GraphEdit~\cite{gllm-gedit} utilizes LLMs to learn complex node relationships within graph data, effectively denoising noisy connections.
LLM4RGNN~\cite{gllm-l4r} is a robust graph structure inference framework, employing LLMs to identify malicious edges and recover missing important edges, thereby reconstructing a robust graph structure.

\section{Conclusion}
In this paper, we introduced a novel framework, \textit{ATAG-LLM}, for conducting black-box graph injection attacks on text-attributed graphs. Our approach addresses the critical challenges of generating effective and interpretable attack samples at the text level by leveraging the capabilities of LLMs to produce node content that balances dissimilarity with contextual relevance. Our framework innovatively employs multiple strategies for LLM prompting, ensuring that the generated text adheres to topic and category constraints, thereby minimizing the detectability of the attack. Additionally, we developed a similarity assessment method to evaluate the effectiveness of node injections, significantly reducing the dependency on surrogate models and lowering the associated costs of model training. Through extensive experiments on two real-world TAG datasets, we demonstrated that \textit{ATAG-LLM} outperforms existing state-of-the-art methods, both at the embedding level and text level, across various scenarios.
Our approach enhances the realism and applicability of graph injection attacks.

\section*{Acknowledgment}

This work was supported in part by the National Natural Science Foundation of China(NO.62272117).

\bibliographystyle{IEEEtran}
\bibliography{refs}

\begin{thebibliography}{10}
\providecommand{\url}[1]{#1}
\csname url@samestyle\endcsname
\providecommand{\newblock}{\relax}
\providecommand{\bibinfo}[2]{#2}
\providecommand{\BIBentrySTDinterwordspacing}{\spaceskip=0pt\relax}
\providecommand{\BIBentryALTinterwordstretchfactor}{4}
\providecommand{\BIBentryALTinterwordspacing}{\spaceskip=\fontdimen2\font plus
\BIBentryALTinterwordstretchfactor\fontdimen3\font minus \fontdimen4\font\relax}
\providecommand{\BIBforeignlanguage}[2]{{%
\expandafter\ifx\csname l@#1\endcsname\relax
\typeout{** WARNING: IEEEtran.bst: No hyphenation pattern has been}%
\typeout{** loaded for the language `#1'. Using the pattern for}%
\typeout{** the default language instead.}%
\else
\language=\csname l@#1\endcsname
\fi
#2}}
\providecommand{\BIBdecl}{\relax}
\BIBdecl

\bibitem{tag1}
H.~Yan, C.~Li, R.~Long, C.~Yan, J.~Zhao, W.~Zhuang, J.~Yin, P.~Zhang, W.~Han, H.~Sun, W.~Deng, Q.~Zhang, L.~Sun, X.~Xie, and S.~Wang, ``A comprehensive study on text-attributed graphs: Benchmarking and rethinking,'' in \emph{Proc. Advances in Neural Information Processing Systems 36: Annual Conference on Neural Information Processing Systems 2023}, New Orleans, USA, 2023.

\bibitem{GAD1}
H.~Dai, H.~Li, T.~Tian, X.~Huang, L.~Wang, J.~Zhu, and L.~Song, ``Adversarial attack on graph structured data,'' in \emph{Proc. 35th International Conference on Machine Learning}, Stockholm, Sweden, Jul. 2018, pp. 1123--1132.

\bibitem{GAD2}
D.~Z{\"{u}}gner, A.~Akbarnejad, and S.~G{\"{u}}nnemann, ``Adversarial attacks on neural networks for graph data,'' in \emph{Proc. 28th International Joint Conference on Artificial Intelligence}, Macao, China, Aug. 2019, pp. 6246--6250.

\bibitem{TAD-Textbugger}
J.~Li, S.~Ji, T.~Du, B.~Li, and T.~Wang, ``Textbugger: Generating adversarial text against real-world applications,'' \emph{arXiv preprint}, vol. arXiv:1812.05271, 2018.

\bibitem{TAD-PETGEN}
B.~He, M.~Ahamad, and S.~Kumar, ``Petgen: Personalized text generation attack on deep sequence embedding-based classification models,'' in \emph{Proc. 27th ACM SIGKDD Conference on Knowledge Discovery and Data Mining}, Virtual Event, Singapore, Aug. 2021, pp. 575--584.

\bibitem{TAD-FastTextDodger}
X.~Hu, G.~Liu, B.~Zheng, L.~Zhao, Q.~Wang, Y.~Zhang, and M.~Du, ``Fasttextdodger: Decision-based adversarial attack against black-box nlp models with extremely high efficiency,'' \emph{IEEE Trans. Inf. Forensics Secur.}, vol.~19, pp. 2398--2411, 2024.

\bibitem{GAD-survey}
L.~Sun, Y.~Dou, C.~Yang, K.~Zhang, J.~Wang, S.~Y. Philip, L.~He, and B.~Li, ``Adversarial attack and defense on graph data: A survey,'' \emph{IEEE Trans. Knowl. Data Eng.}, vol.~35, no.~8, pp. 7693--7711, 2022.

\bibitem{GAD3}
A.~Bojchevski and S.~G{\"{u}}nnemann, ``Adversarial attacks on node embeddings via graph poisoning,'' in \emph{Proc. 36th International Conference on Machine Learning}, Long Beach, CA, USA, Jun. 2019, pp. 695--704.

\bibitem{gia-g2a2c}
M.~Ju, Y.~Fan, C.~Zhang, and Y.~Ye, ``Let graph be the go board: Gradient-free node injection attack for graph neural networks via reinforcement learning,'' in \emph{Proc. Thirty-Seventh AAAI Conference on Artificial Intelligence}, Washington, DC, USA, Feb. 2023, pp. 4383--4390.

\bibitem{gia-tdgia}
X.~Zou, Q.~Zheng, Y.~Dong, X.~Guan, E.~Kharlamov, J.~Lu, and J.~Tang, ``Tdgia: Effective injection attacks on graph neural networks,'' in \emph{Proc. 27th ACM SIGKDD Conference on Knowledge Discovery and Data Mining}, Virtual Event, Singapore, Aug. 2021, pp. 2461--2471.

\bibitem{gia-hao}
Y.~Chen, H.~Yang, Y.~Zhang, K.~Ma, T.~Liu, B.~Han, and J.~Cheng, ``Understanding and improving graph injection attack by promoting unnoticeability,'' in \emph{Proc. The Tenth International Conference on Learning Representations}, Apr. 2022.

\bibitem{gia-camouflage}
S.~Tao, Q.~Cao, H.~Shen, Y.~Wu, L.~Hou, F.~Sun, and X.~Cheng, ``Adversarial camouflage for node injection attack on graphs,'' \emph{Inf. Sci.}, vol. 649, p. 119611, 2023.

\bibitem{gia-single}
S.~Tao, Q.~Cao, H.~Shen, J.~Huang, Y.~Wu, and X.~Cheng, ``Single node injection attack against graph neural networks,'' in \emph{Proc. 30th ACM International Conference on Information and Knowledge Management}, Virtual Event, Queensland, Australia, Nov. 2021, pp. 1794--1803.

\bibitem{gia-scalable}
J.~Wang, M.~Luo, F.~Suya, J.~Li, Z.~Yang, and Q.~Zheng, ``Scalable attack on graph data by injecting vicious nodes,'' \emph{Data Min. Knowl. Discov.}, vol.~34, no.~5, pp. 1363--1389, 2020.

\bibitem{gia-text}
R.~Lei, Y.~Hu, Y.~Ren, and Z.~Wei, ``Intruding with words: Towards understanding graph injection attacks at the text level,'' \emph{CoRR}, vol. abs/2405.16405, 2024.

\bibitem{heter1}
J.~Zhu, J.~Jin, D.~Loveland, M.~T. Schaub, and D.~Koutra, ``How does heterophily impact the robustness of graph neural networks?: Theoretical connections and practical implications,'' in \emph{Proc. 28th ACM SIGKDD Conference on Knowledge Discovery and Data Mining}, Washington, DC, USA, Aug. 2022, pp. 2637--2647.

\bibitem{gnnex}
Z.~Ying, D.~Bourgeois, J.~You, M.~Zitnik, and J.~Leskovec, ``Gnnexplainer: Generating explanations for graph neural networks,'' in \emph{Proc. Advances in Neural Information Processing Systems 32}, 2019, pp. 9240--9251.

\bibitem{gcn}
T.~N. Kipf and M.~Welling, ``Semi-supervised classification with graph convolutional networks,'' in \emph{Proc. Intl. Conf. Learning Representations}, 2017.

\bibitem{gia-nipa}
Y.~Sun, S.~Wang, X.~Tang, T.~Hsieh, and V.~G. Honavar, ``Node injection attacks on graphs via reinforcement learning,'' \emph{CoRR}, vol. abs/1909.06543, 2019.

\bibitem{gia-mia}
Y.~Luo, Y.~Li, D.~Wen, and L.~Lan, ``Message injection attack on rumor detection under the black-box evasion setting using large language model,'' in \emph{Proc. ACM Web Conference 2024}, Singapore, May 2024, pp. 4512--4522.

\bibitem{gllm-survey}
Y.~Li, Z.~Li, P.~Wang, J.~Li, X.~Sun, H.~Cheng, and J.~X. Yu, ``A survey of graph meets large language model: Progress and future directions,'' in \emph{Proc. Thirty-Third International Joint Conference on Artificial Intelligence (IJCAI)}.\hskip 1em plus 0.5em minus 0.4em\relax Jeju, South Korea: ijcai.org, Aug. 2024, pp. 8123--8131.

\bibitem{gllm-robust}
K.~Guo, Z.~Liu, Z.~Chen, H.~Wen, W.~Jin, J.~Tang, and Y.~Chang, ``Learning on graphs with large language models(llms): A deep dive into model robustness,'' \emph{CoRR}, vol. abs/2407.12068, 2024.

\bibitem{gllm-gedit}
Z.~Guo, L.~Xia, Y.~Yu, Y.~Wang, Z.~Yang, W.~Wei, L.~Pang, T.~Chua, and C.~Huang, ``Graphedit: Large language models for graph structure learning,'' \emph{CoRR}, vol. abs/2402.15183, 2024.

\bibitem{gllm-l4r}
Z.~Zhang, X.~Wang, H.~Zhou, Y.~Yu, M.~Zhang, C.~Yang, and C.~Shi, ``Can large language models improve the adversarial robustness of graph neural networks?'' in \emph{Proc. 31st ACM SIGKDD Conference on Knowledge Discovery and Data Mining}, Toronto, ON, Canada, Aug. 2025, pp. 2008--2019.

\end{thebibliography}

\end{document}